\documentclass[11pt]{article}
\usepackage{acl2016}
\usepackage{times}
\usepackage{url}
\usepackage{latexsym}
\usepackage{graphicx}

\newcounter{examplecounter}

\aclfinalcopy 

\title{Measuring the State of the Art of Automated Pathway Curation Using Graph Algorithms - A Case Study of the mTOR Pathway}

\author{Michael Spranger\\
	Sony Computer Science\\ 
	Laboratories Inc.\\
	Tokyo, Japan\\
  {\tt michael.spranger}\\{\tt @gmail.com} \\\And
  Sucheendra K. Palaniappan \\
 INRIA, \\ Campus de Beaulieu,\\ 
 Rennes, France \\
  {\tt sucheendra.palaniappan}\\{\tt @inria.fr} \\\And
  Samik Ghosh \\
	 The Systems Biology Institute, \\
	 Minato-ku, \\
	 Tokyo, Japan\\
  {\tt ghosh@sbi.jp} \\}

\date{}

\begin{document}
\maketitle
\begin{abstract}
This paper evaluates the difference between human pathway curation and current NLP systems. We propose graph analysis methods for quantifying  the gap between human curated pathway maps and the output of state-of-the-art automatic NLP systems. Evaluation is performed on the popular mTOR pathway. Based on analyzing where current systems perform well and where they fail, we identify possible avenues for progress. 
\end{abstract}

\section{Introduction}
Biological pathways encode sequences of biological 
reactions, such as phosphorylation, activations etc, involving various biological 
species, such as genes, proteins etc., in response to certain stimuli or spontaneous at times
\cite{aldridge2006physicochemical,kitano2002computational}.
Studying and analyzing pathways is crucial to understanding biological systems
and for the development of effective disease treatments and drugs
\cite{creixell2015pathway,khatri2012ten}. There have been numerous efforts to 
reconstruct detailed process-based and disease level pathway maps such as 
Parkinson disease map \cite{fujita2014integrating}, Alzheimers disease Map \cite{mizuno2012alzpathway}, mTOR pathway
 Map \cite{caron2010comprehensive}, and the TLR pathway map \cite{oda2006comprehensive}). 
Traditionally,  these maps are constructed and curated by expert pathway curators who 
 manually read numerous biomedical documents, comprehend and assimilate the knowledge 
 in them and construct the pathway. 

\begin{figure}
\begin{center}
\includegraphics[width=0.65\columnwidth]{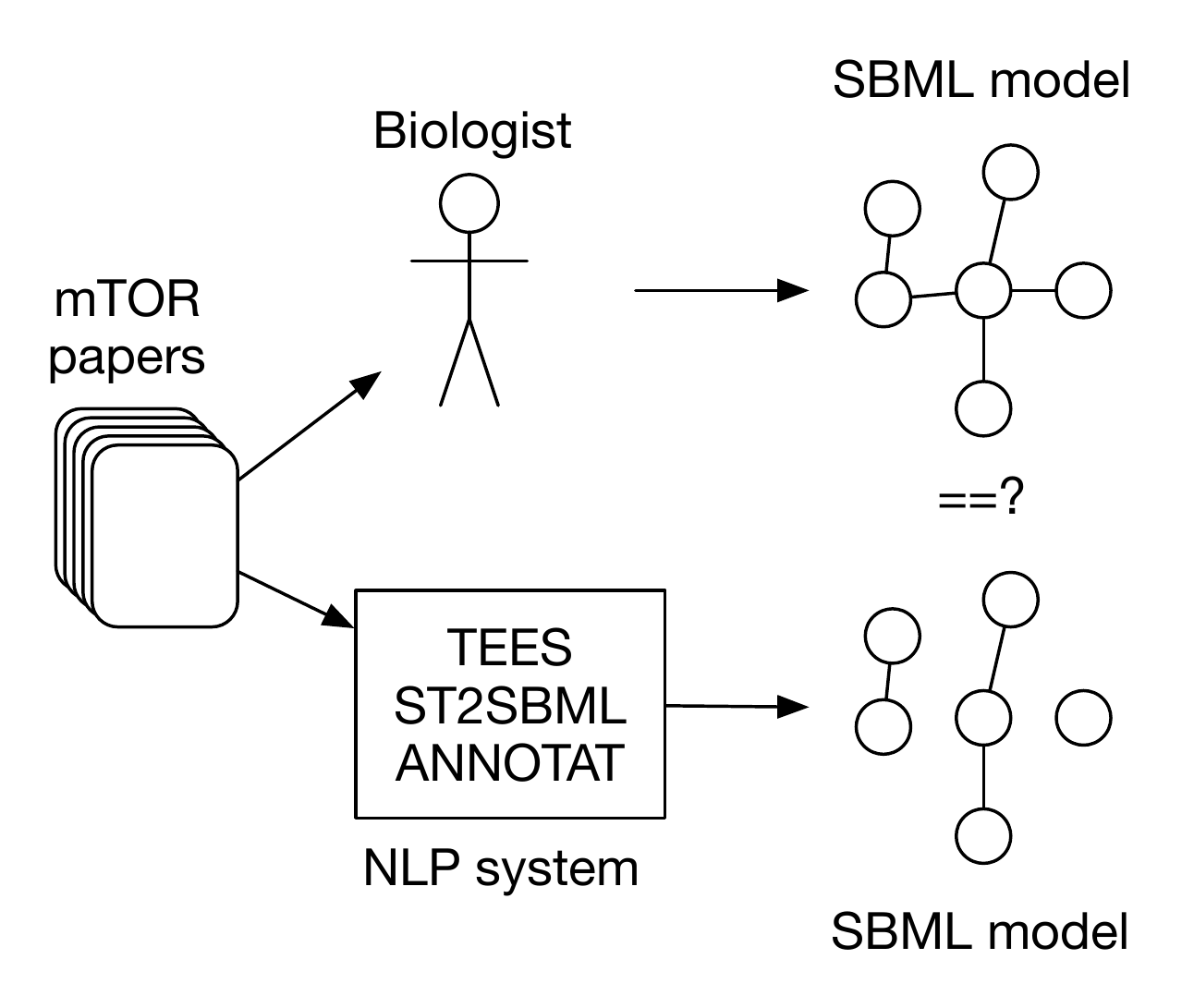}
\end{center}
\caption{Comparing human pathway curation to NLP extraction.}
\label{f:systems}
\end{figure}
	
Manual curation of pathways is rather challenging given the ever increasing barrage 
of scientific publications. It is basically common place in this community that 
manual curation is not sufficient \cite{baumgartner2007manual}.
Consequently, \emph{Automated Pathway Curation} has been an active area of research - 
particularly in the BioNLP community \cite{miwa2012extracting,Valenzuela+:2015aa}. It is also
the goal of large scale research efforts such as DARPA's Big Mechanism Project \cite{bigmechanism}.

NLP systems have shown to perform well in BioNLP competitions \cite{sharedtask2013,ohta2013overview,Ananiadou2010}, 
but so far we do not have systems that automatically assemble and curate pathways of the scope and complexity of, for  example, the
mTOR pathway. This paper investigates why this is the case. We measure the state of the art by 
closing the gap between NLP representations and biological networks, then we apply graph theory
and in particular graph matching to quantify how much overlap there is between the NLP output
and the information that humans assemble (see also Figure \ref{f:systems}). The evaluation is performed
on the popular mTOR pathway. 

This paper starts by introducing our approach, followed by a description of data sets and evaluation
results. We conclude by discussing where current system seem to fail and how to make progress.

\begin{figure}[t]
\begin{center}
\includegraphics[width=0.7\columnwidth]{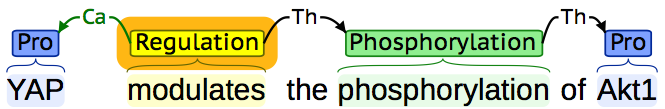}
\end{center}
\caption{Example sentence with NLP event representations extracted.}
\label{f:event}
\begin{center}
\includegraphics[width=0.6\columnwidth]{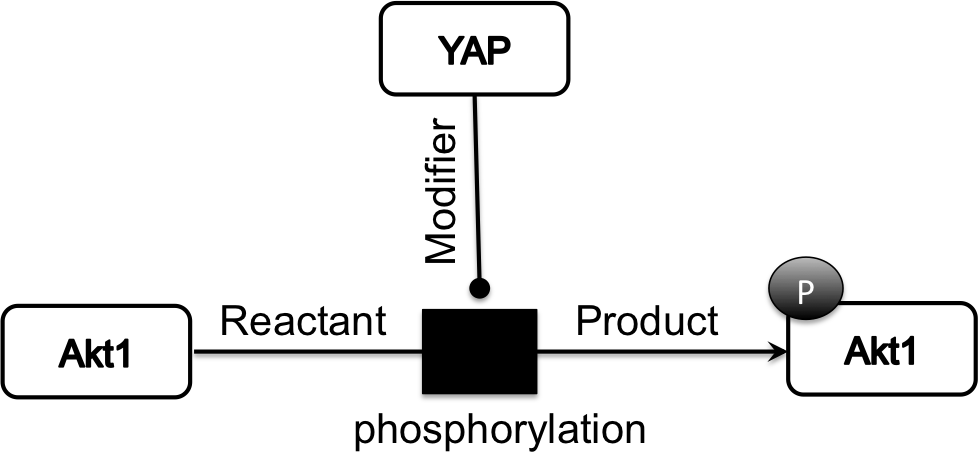}
\end{center}
\caption{Phosphorylation reaction.}
\label{f:reaction}
\end{figure}

\section{Bridging the Gap}
In this paper, we close the representational gaps between current NLP systems 
and human-generated pathways, measure the overlap and analyze possible shortcomings of current systems.
Evaluation is performed on the popular, hand-curated mTOR pathway map \cite{caron2010comprehensive}.
Experts have curated and assembled the information from 522 papers into one large map using CellDesigner 
\cite{funahashi2008celldesigner} - a software
for modeling but also executing mechanistic models of pathways. CellDesigner represents information 
using a heavily customized XML-based SBML format  \cite{hucka2003systems}. 

mTOR has been published along with a list of the 522 papers used to build the map. This allows us
to treat the same papers with state-of-the-art NLP extraction systems. Here we used one of the most
successful NLP systems around - the TURKU event extractions system \cite[TEES]{bjorne2014biomedical}. 
TEES has won 1st place in BioNLP 2009 ST, 2011 ST and DDI 2011 \cite{bjorne2012university}. 
The system integrates various NLP techniques to extract events from text. Processing roughly proceeds as follows 1) A number 
of external tools detect protein names and parse the sentences. 2) The event detector detects 
trigger words such as verbs, which is followed by detection of interactions. 3) Complex 
events are constructed. 4) The system detects modifiers such as negation and speculation.

NLP systems typically operate on something called the standoff format.
From a sentence such as in Figure \ref{f:event}, standoff containing
\emph{entities} and \emph{events} will be extracted. These in principle correspond to 
biological species and reactions. We translate the NLP representation into SBML
pathways and perform additional annotation \cite{spranger2015extracting}
of species and reactions. For the sentence in Figure \ref{f:event}, the extracted SBML 
is visualized in Figure \ref{t:reactions}.

\subsection*{Datasets}

We compared 3 different sets of data all related to mTOR pathway.

\paragraph{MTOR-HMN} is a mTOR pathway map manually constructed
by human expert pathway curators.
\cite{caron2010comprehensive}. The pathway is encoded in a dialect of SBML 
used by CellDesigner  \cite{funahashi2008celldesigner}. We convert the CellDesigner 
format into pure SBML and annotate reactions and species further by automatically
assigning reaction types and gene/protein identifers (see description below).

\paragraph{MTOR-ANN} consists of 57 abstracts of scientific papers from Pubmed
related to the mTORpathway map. The data set was \emph{human-annotated} for 
NLP system training \cite[Corpus annotations (c) GENIA 
Project\footnote{http://nactem.ac.uk/GENIA/current/Other-corpora/mTOR-Pathway-Events/}]{ohta2011pathways}. 
This corpus gives an idea of the potential performance of a machine with human-level
NLP extraction capabilities. Annotated NLP entities and events were used to create SBML 
representations and further annotated using various tools (discussed below).

\paragraph{MTOR-NLP} consists of 522 full text papers mentioned in the mTOR pathway map. 
Paper pdfs were downloaded automatically and translated into raw txt files using CERMINE \cite{tkaczyk2015cermine}. We managed to extract text from 501 papers. The 501 papers were processed using the Turku Event Extraction System
mentioned earlier. From the extracted NLP events we created SBML representations 
of pathway maps for each text using \cite{spranger2015extracting}. The SBML
was further annotated using various tools (discussed below) and,
finally, loaded into a single pathway map.

Notice that MTOR-ANN and MTOR-NLP are different in how they are constructed and 
consequently what kind of conclusion we can draw from them. MTOR-ANN is a human-annotated dataset
which contains much less data than MTOR-NLP. However, because it is human-annotated
it allows us to evaluate a human-level performance extraction systems. So we cannot
expect that MTOR-ANN is able to reconstruct everything in MTOR-HMN (recall). However, 
as we will argue in this paper, we might expect that what is extracted in MTOR-ANN does 
occur in MTOR-HMN (high precision).

The following table shows number of species, reactions and edges between them
for the different datasets.
\begin{footnotesize}
\begin{center}
	\begin{tabular}{ |p{2.5cm}|r|r|r|}
		\hline
		Dataset & \# species & \# reactions & \# edges \\ 
		\hline
		MTOR-HMN & 2242  & 777 & 2457\\
		\hline
		MTOR-ANN & 2457 & 857 & 2343\\
		\hline
		MTOR-NLP & 292049 & 100130 & 203042\\
		\hline
	\end{tabular}
\end{center}
\end{footnotesize}

\subsection*{Annotation}

\paragraph{Annotation SBO}
Reactions in datasets MTOR-HMN, MTOR-ANN and MTOR-NLP were automatically \emph{annotated} 
using Systems Biology Ontology (SBO) \cite{le2006model} and Gene Ontology (GO) terms. SBO provides a class hierarchy 
of reactions. Reactions can be of a certain type. For instance, NLP systems often identify
regulation events. Regulation reactions form a hierarchy. For instance, positive regulation
is a subclass of regulation reactions. Phosphorylation reactions are a subclass of conversion reactions.

All reactions in MTOR-HMN, MTOR-ANN, and MTOR-NLP are annotated using SBO/GO (coverage 100\%). 
SBO/GO annotations are computed using different approaches. For MTOR-ANN and MTOR-NLP
we used an automated annotation system that is also used to convert NLP event representations to SBML \cite{spranger2015extracting}. 
For MTOR-HMN, we used annotations provided by humans extended by automatic annotations.
Automatic annotations were deduced by examining the reactants and products of reactions. For example, 
if a phosphoryl group is added the reaction is annotated using the SBO term for phosphorylation.
Notice, in MTOR-HMN each reaction can be annotated with multiple SBO/GO terms. For instance,
a single reaction can be annotated as phosphorylation and activation. This is not the case
for MTOR-ANN and MTOR-NLP where each reaction corresponds to exactly one SBO/GO term.

\begin{table}
\begin{footnotesize}
\begin{center}
	\begin{tabular}{ |p{2.5cm}|p{1cm}|p{1cm}|p{1cm}|}
		\hline
		& MTOR-HMN & MTOR-ANN & MTOR-NLP\\\hline
		activation & 72 & 104 & 16485 \\\hline
		association & 210 & 204 & 21055 \\\hline
		conversion & 171	 & 0 & 0 \\\hline
		deacetylation & 1 & 0 & 0 \\\hline
		dephosphorylation & 28 & 14 & 0 \\\hline
		deubiquitination & 13 & 0 &  0\\\hline
		dissociation & 43 & 55 & 0 \\\hline
		gene expression & 4  & 40 & 18810 \\\hline
		localization & 0 & 16  & 474 \\\hline
		negative regulation & 33 & 99 & 10723 \\\hline
		phosphorylation & 85 & 241 & 25406 \\\hline
		protein catabolism & 24 & 18 & 1080 \\\hline
		regulation & 0 & 0 & 4832 \\\hline
		transcription & 78 & 8 & 1265 \\\hline
		translation & 23 & 1 & 0 \\\hline
		transport & 87 & 53 & 0 \\\hline
		ubiquitination & 13 & 4 & 0 \\\hline

	\end{tabular}
	\label{t:reactions}
	\caption{Reaction types extracted and annotated for various data sets. All reactions are annotated
	with their most specific type. Numbers are non-cumulative. For instance, the 171 conversion operations
	in MTOR-HMN	are only annotated with the general conversion (SBO:182)  and not more specific reaction types.}
\end{center}
\end{footnotesize}
\end{table}

\paragraph{Annotation Entrez Gene}
Species in all three datasets were annotated using the
gene/protein named entity recognition and normalization software 
GNAT  \cite{hakenberg2011gnat} - a publicly available
gene/protein normalization tool. GNAT returns a set of Entrez Gene 
identifiers \cite{maglott2005entrez} for each input string. Species were annotated using all
returned Entrez Gene identifiers for a particular species (organism human).
We call the set of Entrez Gene identifiers returned by GNAT for each species
 \emph{Entrez Gene signature}.

\begin{footnotesize}
\begin{center}
	\begin{tabular}{ |p{2cm}|r|r|r|}
		\hline
		 & \# species & coverage & \# Entrez ids\\ 
		\hline
		MTOR-HMN & 2242  & 90\% & 538 \\
		\hline
		MTOR-ANN & 2457 & 87\%& 317 \\
		\hline
		MTOR-NLP & 292049 & 83\% & 4194 \\
		\hline
	\end{tabular}
\end{center}
\end{footnotesize}

\section{Species}

Pathways contain many references to the same protein or 
gene. We measured the number of unique 
genes and proteins in each dataset using various ways of 
identifying (normalizing) genes and proteins in a particular dataset.

\begin{footnotesize}
\begin{center}
\begin{tabular}{ |p{2.4cm}|r|r|r|}
\hline
& \multicolumn{1}{p{1.1cm}|}{MTOR-HMN} 
& \multicolumn{1}{p{1.1cm}|}{MTOR-ANN} 
& \multicolumn{1}{p{1.1cm}|}{MTOR-NLP} \\\hline
\# species & 2242 & 2457 & 291218\\\hline
\# names & 582 & 359 & 27928 \\\hline
\# appr names & 568 & 316 & 4517 \\\hline
\# Entrez signatures & 443 & 201 & 6220 \\\hline
\end{tabular}
\end{center}
\end{footnotesize}

The first row repeats the number of species per data 
set. The second row condenses the species names by removing
prefixes such as ``phosphorylated'' and other adjectives irrelevant
for determining the actual biological entity. The third row shows
what happens when we reduce the names further by using a
Levenshtein-based string distance with a cutoff point of 90.
The last row measures how many different unique Entrez Gene id 
signatures there are. Each species is annotated with a set
of Entrez Gene ids. The set of the Entrez Gene identifiers for
each species is taken as a signature. 

The numbers show the degree of redundancy or reuse of species
within each pathway. They also suggest that there are far more 
species implicated in MTOR-NLP than there are in MTOR-HMN.
In other words, human annotators of mTOR have selected 
568 species and not the 4517 found by the NLP systems (approx names). 

\paragraph{Unique Species Overlap} To better understand species identification 
we can measure the overlap
of MTOR-ANN and MTOR-NLP with MTOR-HMN based on the unique
species. Here we consider names equal (\emph{nmeq}), names approximately 
equal (\emph{appeq}), Entrez Gene id signature equal (\emph{enteq})
and Entrez Gene id signature overlap (\emph{entov}).
The focus is on unique items.

\begin{footnotesize}
\begin{center}
	\begin{tabular}{ |p{3cm}|r|r|r|}
		\hline
		 & precision & recall & f-score\\\hline
		 \multicolumn{4}{|l|}{MTOR-HMN/MTOR-ANN} \\\hline
		nmeq & 20.89 & 12.89 & 15.94 \\\hline
		appeq & 27.30 & 15.64 & 19.88 \\\hline
		enteq & 45.27 & 20.54 & 28.26 \\\hline
		entov & 83.08 & 55.53 & 66.57 \\\hline\hline
		 \multicolumn{4}{|l|}{MTOR-HMN/MTOR-NLP} \\\hline
		nmeq & 0.96 & 45.88 & 1.87 \\\hline
		appeq & 1.59 & 51.20 & 3.08 \\\hline
		enteq & 4.60 & 64.56 & 8.58 \\\hline
		entov & 58.04 & 99.55 & 73.33 \\\hline
	\end{tabular}
\end{center}
\end{footnotesize}

The rows \emph{nmeq} show precision and recall for unique species names in MTOR-NLP with respect
to MTOR-HMN. Precision is low - meaning that only
a small percentage of unique species names in MTOR-NLP actually 
appear in MTOR-HMN. On the other hand, recall is higher. This shows 
that the few correctly identified species in MTOR-NLP overlap with 
large parts of MTOR-HMN species. Less than a percent of unique species names in 
MTOR-NLP cover 46\% of species in MTOR-HMN.
What is interesting is that MTOR-ANN does not fair
too great on precision either. 79\% of the unique annotated names do not
appear in MTOR-HMN. Especially the annotated version dataset MTOR-ANN,
lets us conclude that many species mentioned 
in papers actually do NOT make it into the pathway or at least not as mentioned
in the papers. These analyses point to the fact that researchers building 
pathways select species. In other words, pathway curation is 
\emph{not just extraction}, but \emph{active selection} and, in fact, 
\emph{identification} of species with proteins and genes known to the scientist.

\paragraph{Complex Species} MTOR-HMN pathway contains a lot 
of complex species - i.e. species that contain other species. There are 351 complex species with a total 
of 1192 total constituents. 16 complexes are part of other complexes. Together this accounts for
more than 70\% of the species in MTOR-HMN. In other words, this is important
information. Both MTOR-NLP and MTOR-ANN do not provide information about
complexes explicitly. However, for this paper complexes are essentially treated 
like any other species.

\section{Reactions}
We first measured how many unique reaction types there are for each of the datasets.

\begin{footnotesize}
\begin{center}
	\begin{tabular}{ |p{2cm}|r|r|r|r|}
		\hline
		& \multicolumn{1}{p{1.4cm}|}{\# reactions} 
			   & \multicolumn{1}{p{1.4cm}|}{\# SBO/GO terms} 
			  & \multicolumn{1}{p{1.4cm}|}{\# SBO/GO signatures} \\\hline
		MTOR-HMN & 777 & 15 & 29 \\\hline
		MTOR-ANN & 857 & 13 & 13 \\\hline
		MTOR-NLP & 100130 & 9 & 9 \\\hline
	\end{tabular}
\end{center}
\end{footnotesize}

MTOR-HMN contains 777 reactions with 12 SBO/GO terms, i.e. reaction types.
MTOR-ANN contains 12 and  MTOR-NLP slightly less.
Each reaction can have multiple SBO/GO terms associated
with it. We call this the SBO/GO signature of a reaction. 
For instance, a particular reaction can be typed as phosphorylation 
and activation. Its signature are then the  SBO/GO terms for these 2 reactions.
The table shows that this actually only happens in MTOR-HMN. Human annotators are free to combine various reactions
 into a single reaction if they see fit. There is no replication of this in
 the automated data.

\paragraph{Unique Reaction Signature Overlap } We then measured how much unique signatures overlap across 
the different datasets. We checked three different measures:
1) \emph{sboeq} requires that both signatures are the same, 2) \emph{sboov}
requires that the intersection of the signatures overlaps - i.e. is not empty - and
3) \emph{sboisa} requires that there is at least one
SBO/GO term in each signature that relate in a is\_a relationship
in the SBO reaction type hierarchy. For instance, if there is a phosphorylation
reaction and a conversion reaction, then \emph{sboisa} will match because
phosphorylation is a subclass of conversion according to the SBO type hierarchy.

\begin{footnotesize}
\begin{center}
	\begin{tabular}{ |p{2cm}|r|r|r|}
		\hline
		& precision & recall & f-score\\\hline
		\multicolumn{4}{|l|}{MTOR-HMN/MTOR-ANN} \\\hline
		sboeq & 69.23 & 31.03 & 42.86  \\\hline
		sboov & 45.51 & 50.19 & 47.74  \\\hline
		sboisa & 92.31 & 93.10 & 92.70  \\\hline
		\multicolumn{4}{|l|}{MTOR-HMN/MTOR-NLP} \\\hline
		sboeq & 55.56 & 17.24 & 26.32  \\\hline
		sboov & 77.78 & 68.97 & 73.11  \\\hline
		sboisa & 88.89 & 79.31 & 83.83  \\\hline
	\end{tabular}
\end{center}
\end{footnotesize}

MTOR-ANN catches 1/3 of the reaction SBO/GO signatures directly 
and up to 93\% when we allow for overlap sbo\_is\_a relationship. MTOR-NLP only 
directly includes 1 out of 5 reaction signatures. However, the overlap is higher
when allowing for reaction SBO/GO signatures to overlap and individual SBO terms
to be in a is\_a relationship.

These results also show that there are reactions in MTOR-NLP and MTOR-ANN
that are not part of MTOR-HMN (see also Table \ref{t:reactions})

From this preliminary data, we can immediately identify an important difference between
human annotation and automated NLP event extraction. Human annotators combine 
multiple reactions into a single reaction representation to condense information. 

\section{Networks - Connectedness}
Ultimately we are interested in networks of reactions
and species. Studying the output of NLP systems it becomes immediately clear
that the result of these systems differs from hand-curated data in an important
aspect: \emph{connectedness}. To show this we measured isolation of species 
and networks (reactions cannot be isolated for structural reasons in SBML). 

\begin{footnotesize}
\begin{center}
	\begin{tabular}{ |l|r|r|r|}
		\hline
		 & \# isolated networks & \# isolated species \\\hline
		MTOR-HMN & 4 & 6 \\\hline
		MTOR-ANN & 475 & 632 \\\hline
		MTOR-NLP & 83,093 & 110,490 \\\hline
	\end{tabular}
\end{center}
\end{footnotesize}

In MTOR-HMN there are 4 separate subgraphs (no connection between them). 3
of them are modeling mistakes by human curators. Basically MTOR-HMN is 
one connected network. On the other hand, MTOR-ANN and MTOR-NLP
consist of numerous unconnected networks. Each of them is quite small as the following
data shows.

We measured min, max, mean and median number of species and reactions 
in each connected component subgraph.

\begin{footnotesize}
\begin{center}
	\begin{tabular}{ |p{2cm}|r|r|r|r|}
		\hline
		dataset & min & mean & median & max \\\hline
		MTOR-ANN & 1 & 3.00 & 1.0 & 24 \\\hline
		MTOR-NLP & 1 & 2.02 & 1.0 & 215 \\\hline
	\end{tabular}
\end{center}
\end{footnotesize}

Results show that subgraphs in MTOR-ANN and MTOR-NLP on average 
contain between 2 and 3 species and reactions. So very often there will 
be a single reaction in a subgraph plus some reactant and maybe a product.
On the other hand MTOR-HMN consists of essentially one large connected
graph. So here is another fundamental difference: human modelers \emph{compose} 
a single large graph, as opposed to just extracting single reactions.

\section{Networks - Overlap}
Arguably the most important question is how much overlap there is between
disconnected reactions extracted by MTOR-ANN/MTOR-NLP with MTOR-HMN. 
For this, we measure subgraph isomorphisms of MTOR-ANN and MTOR-NLP
subgraphs with the MTOR-HMN graph. We measured \emph{max} overlap
and allow multiple hits for each subgraph from MTOR-ANN and MTOR-NLP 
with parts of MTOR-HMN. We compare different 
strategies for node (species and reactions) and edge matching.

\paragraph{Species matching} We investigated name matches (\emph{nmeq}), approximate name matches
	(\emph{appeq}), Entrez Gene signature equal (\emph{enteq}) and Entrez Gene signature overlaps (\emph{entov}) and combinations
	thereof. For example, \emph{appeq/enteq} matches two species if either their names match approximately OR their Entrez Gene signatures are equal. \emph{appeq/entov} matches two species if their names match approximately OR their Entrez Gene signatures overlap.
	Since there is no information on complexes in MTOR-ANN/MTOR-NLP, we
	also allowed matches not only on the complex itself but also on its constituents (\emph{wc}). So a link 
	present in MTOR-NLP between some protein and its phosphorylated version, will match
	if a link is present in a complex that contains that protein in MTOR-HMN.
	
\paragraph{Reaction matching} Reaction matching relies on SBO/GO signatures. We checked 
with signatures equal (\emph{sboeq}), signatures overlapping overlap (\emph{sboov}) and signatures overlapping with
individual SBO terms in is\_a relationship (\emph{sboisa}).
	
\paragraph{Edge matching} We only allowed strict edge matching. So if an edge marks
	a reactant, then it has to be a reactant in MTOR-HMN. Same holds for product and modifier.

\begin{figure}
\begin{center}
\includegraphics[width=0.8\columnwidth]{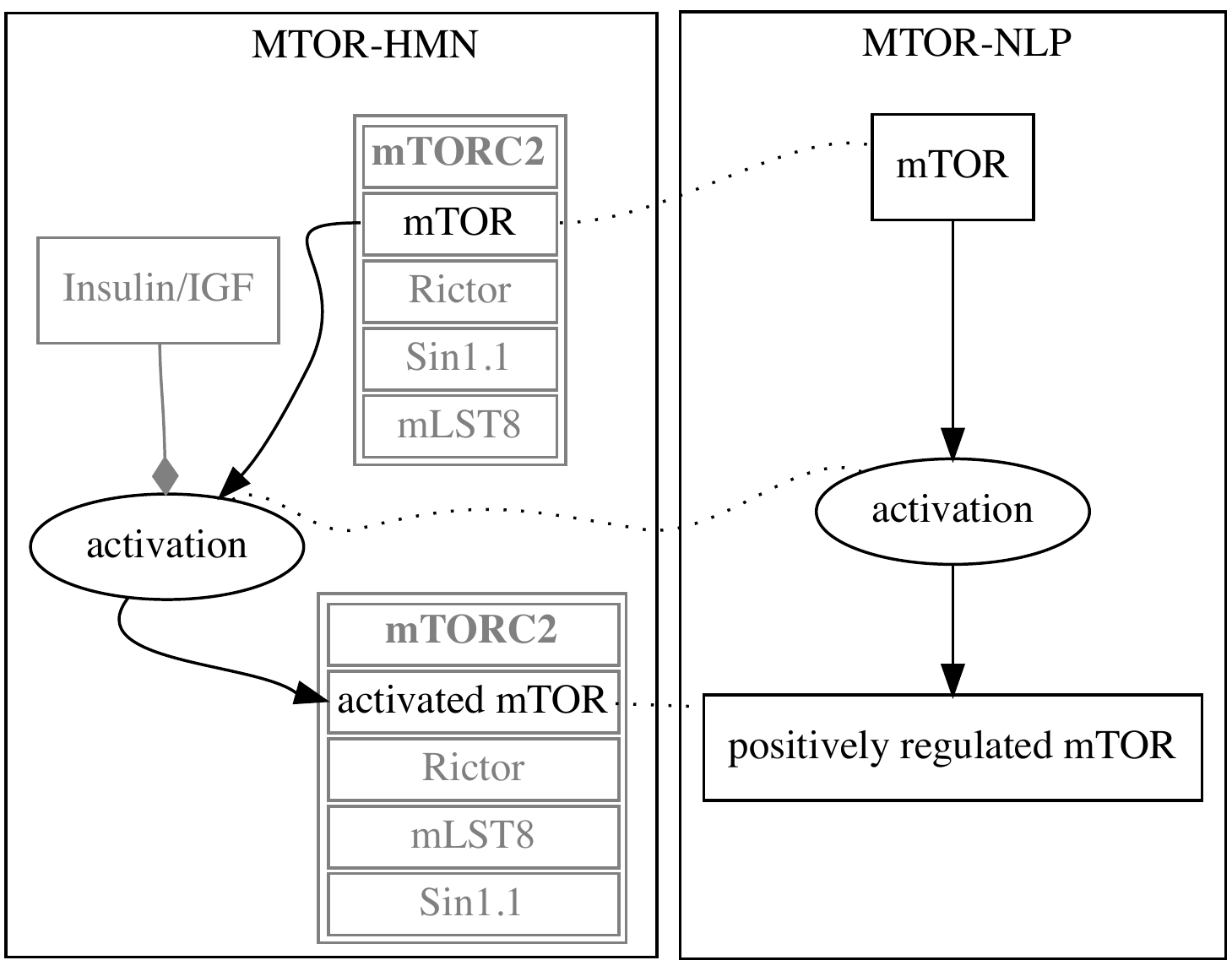}
\caption{Example of a successful match (\emph{nmeq, sboeq}). Black - matched nodes and edges, grey not mached context. Insulin/IGF is
a modifier of this reaction. It is not captured by MTOR-NLP. Modifiers are less frequently detected than reactants and 
products.}
\label{f:graph-1}
\vspace{.3cm}
\includegraphics[width=0.8\columnwidth]{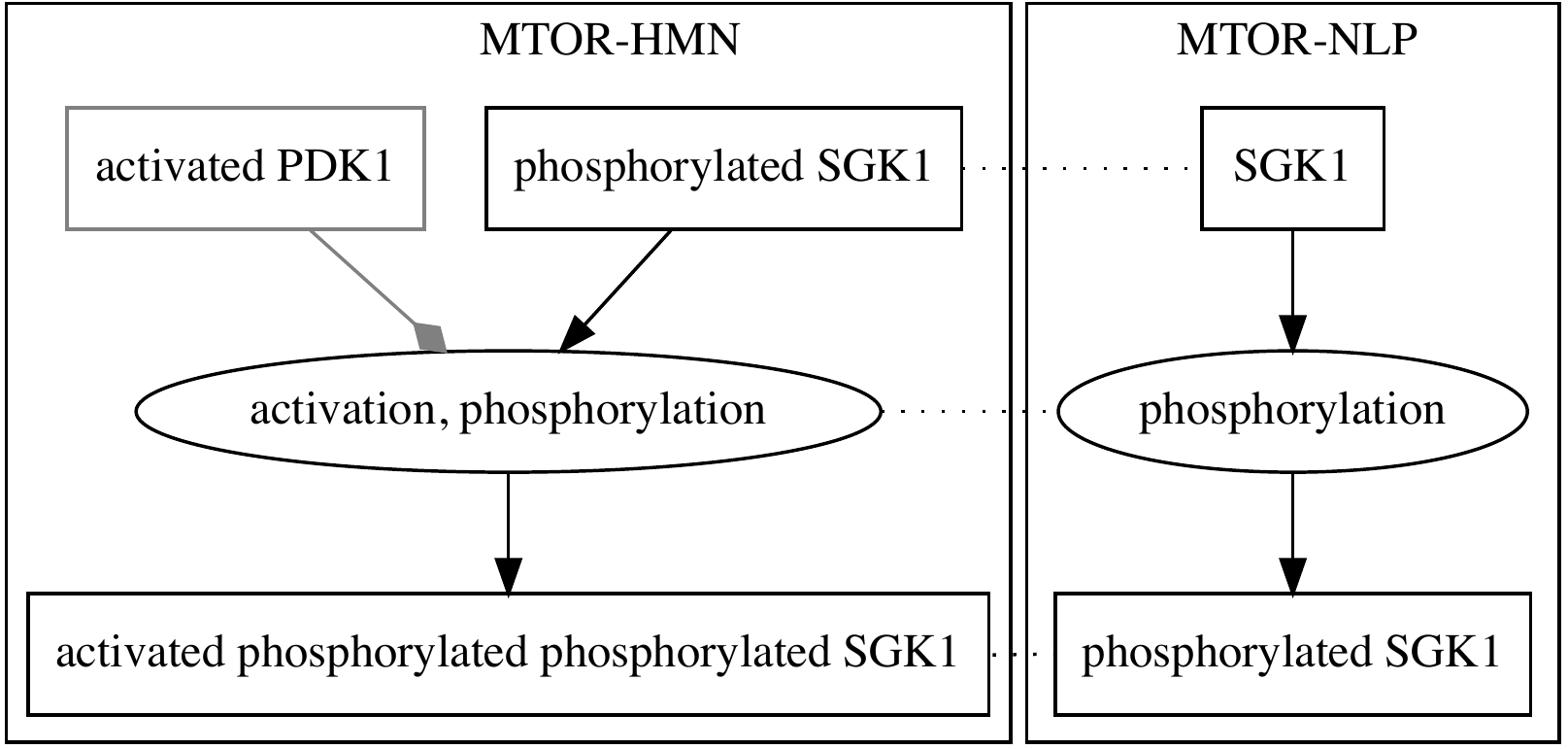}
\caption{Example of a successful match (\emph{appeq, sbois}) with a reaction that has
multiple reaction types.}
\label{f:graph-2}
\vspace{.3cm}
\includegraphics[width=0.8\columnwidth]{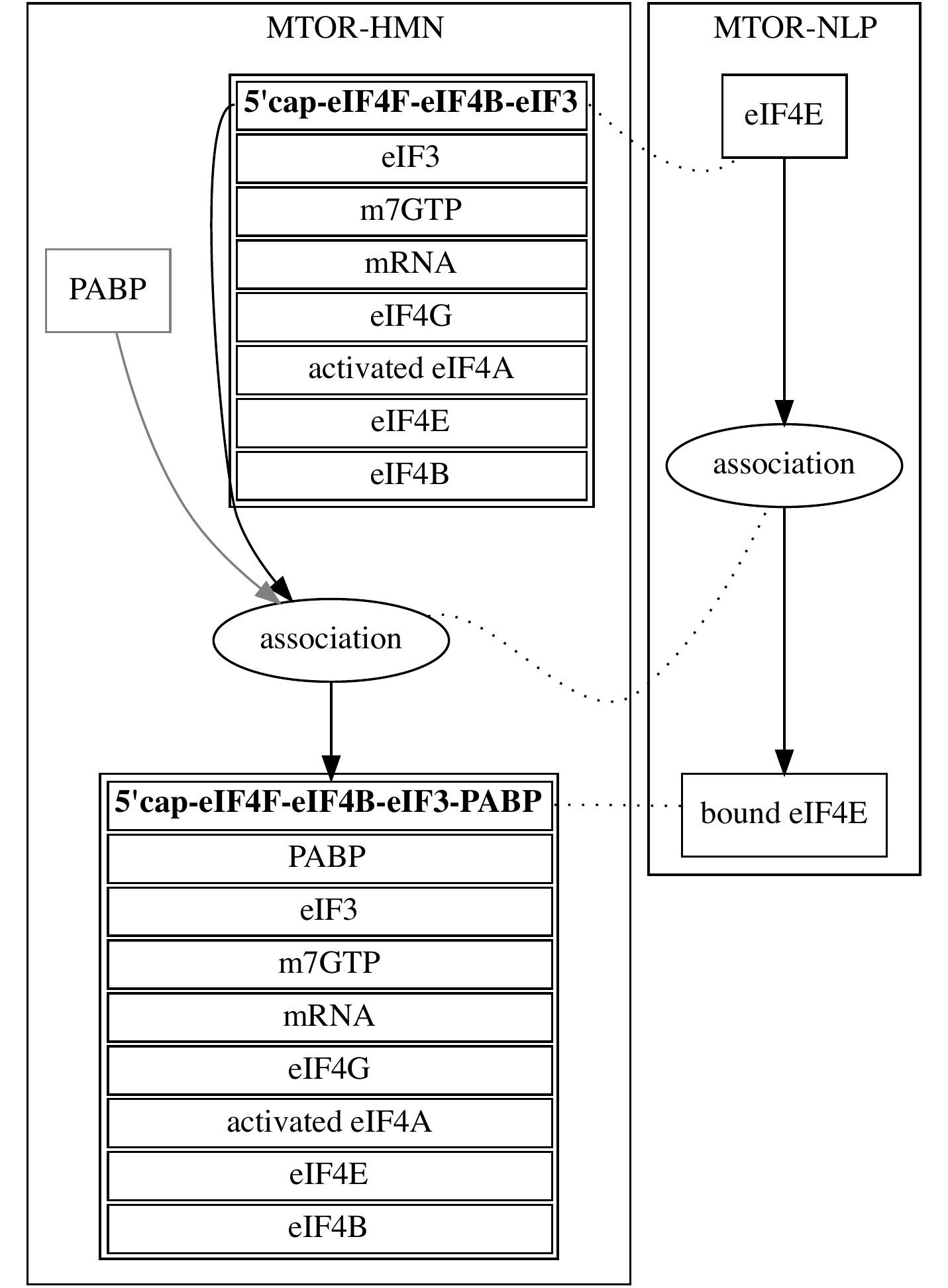}
\caption{Example of a successful match (\emph{appeq/wc, sboeq}) but ultimately incorrect mapping.
It is not eIF4E that gets bound but the whole complex of 5'cap-eIF4F-eIF4B-eIF3 
that includes eIF4E.
}
\label{f:graph-3}
\end{center}
\end{figure}

The final point to note for the results of matching is that we removed isolated nodes (which are always species) 
from MTOR-ANN and MTOR-NLP, because here we are really interested in graph structure. 

\paragraph{Network overlap results} Table \ref{t:network} shows precision and recall for max 
overlap of different matching strategies (see also Figures \ref{f:graph-1} to \ref{f:graph-3}). The table shows results for MTOR-ANN and MTOR-NLP
successively. In general the first rows (\emph{nmeq, sboeq}) represent very strict matching strategies.
The last row (\emph{appeq/entov/wc, sbois}) shows results for the most ``relaxed'' strategy.

Let us first analyze the performance of MTOR-NLP. The automated NLP system is able to retrieve
roughly 9\% of all edges given the strictest matching strategy. This means that 1 in 10 edges 
in the NLP extracted dataset actually appears as is in the human curated data (MTOR-HMN). Also,
if we look at the most relaxed matching strategy \emph{appeq/entov/wc, sbois}, we find that
roughly 2 of 3 edges and 3 of 4 nodes (species and reactions) in the human curated MTOR 
have something to do with the NLP extracted data. In particular, the conversion and regulation reactions
play a part in the 20 percentage points jump from 45.59 to 65.04 for edges from \emph{appeq/entov/wc, sboov} 
to \emph{appeq/entov/wc, sboisa} matching. Conversion and regulation are super classes for a whole range of 
reactions (conversion: phosphorylation etc; regulation: activation, inactivation etc).

Matching strategies that allow for matching complex constituents always have a higher recall and precision
performance than their non constituent matching counterparts. For instance, \emph{nmeq, sboeq} matches 
almost 20 percentage points less edges than \emph{nmeq/wc, sboeq} (MTOR-HMN/MTOR-NLP).
This increase in performance of constituent matching points to the fact that human modelers
often attribute reactions to the whole complex. For instance, a phosphorylation may be acting
on a constituent of a complex but the human modeler chooses to connect the reaction
with the whole complex. These matching strategies do account for that and therefore
are able to improve the numbers (in some cases) considerably.

Reactions in MTOR-HMN are sometimes incorporating various reaction types.
In MTOR-ANN and MTOR-NLP, on the other hand, each reaction only has a single type. Reaction matching
strategies \emph{sboov} and \emph{sboisa} account for that by looking at overlaps. This means
that reactions in MTOR-ANN and MTOR-NLP will match with a reaction MTOR-HMN if the
reaction type signatures intersection is not empty. In reality this means that the reaction in MTOR-ANN or
MTOR-NLP has to be an element of the reaction in MTOR-HMN.

Lastly, let us take a look at MTOR-HMN/MTOR-ANN. MTOR-ANN contains
much less data than MTOR-NLP but the reason we include it here is because
MTOR-ANN consists of human annotated data. It therefore gives an idea
about the limits of the annotation data and the limits of human annotation.
If all of the problems discussed so far are purely a problem of the NLP system,
then MTOR-ANN should do better than MTOR-NLP in terms of precision but not
in terms of recall. Recall will be low because the MTOR-ANN consists of less data. However, 
we would expect high precision numbers. Interestingly, data shows that even for NLP-ANN 
precision is low. With relaxed matching strategies \emph{appeq/enteq/wc, sboisa} and 
\emph{appeq/entov/wc, sboisa}, we see some substantial recall 20\% (remember NLP-ANN 
is only abstracts). Nevertheless precision for edges is only 1 in 10 and for nodes about the same. 

\begin{table}
\begin{footnotesize}
	\begin{center}
		\begin{tabular}{ |p{2.95cm}|r|r|r|r|}
			\hline
			\multicolumn{5}{|l|}{MTOR-HMN/MTOR-ANN} \\\hline
			&  \multicolumn{2}{c|}{nodes} & \multicolumn{2}{|c|}{edges} \\\hline
			& prec & rec & prec & rec \\\hline
			nmeq, sboeq  &  1.22 & 1.93  & 0.94 & 1.30 \\\hline
			nmeq, sboov  &  1.52 & 2.72  & 1.15 & 1.91 \\\hline 
			nmeq, sboisa  &  3.15 & 4.00  & 2.43 & 2.65\\\hline 
			nmeq/wc, sboeq &  3.48 & 8.60  & 2.77 & 6.76\\\hline
			nmeq/wc, sboov &  3.78 & 9.34  & 2.99 & 7.45 \\\hline 
			nmeq/wc, sboisa &  5.59 & 12.21  & 4.44 & 9.73 \\\hline
			appeq, sboeq  &  1.44 & 2.22  & 1.11 & 1.47 \\\hline
			appeq, sboov  &  1.81 & 3.11  & 1.37 & 2.12 \\\hline 
			appeq, sboisa  &  3.93 & 4.50  & 2.99 & 2.89\\\hline 
			appeq/wc, sboeq &  3.85 & 8.90  & 3.07 & 7.04\\\hline
			appeq/wc, sboov &  4.22 & 9.74  & 3.33 & 7.77\\\hline 
			appeq/wc, sboisa &  6.67 & 12.85  & 5.25 & 10.22 \\\hline 
			appeq/enteq, sboeq  &  2.74 & 3.02  & 2.13 & 1.95 \\\hline
			appeq/enteq, sboov  &  3.19 & 3.86  & 2.43 & 2.65 \\\hline 
			appeq/enteq, sboisa  &  5.81 & 5.78  & 4.48 & 3.74\\\hline 
			appeq/enteq/wc, sboeq &  9.78 & 13.69  & 8.15 & 10.99\\\hline
			appeq/enteq/wc, sboov &  10.48 & 15.37  & 8.66 & 12.54\\\hline 
			appeq/enteq/wc, sboisa & 14.67 & 23.88  & 11.95 & 19.90 \\\hline 
			appeq/entov, sboeq  &  8.85 & 10.33  & 7.34 & 7.41 \\\hline
			appeq/entov, sboov  &  9.41 & 12.01  & 7.73 & 8.79 \\\hline 
			appeq/entov, sboisa  &  13.59 & 19.53  & 11.01 & 14.73\\\hline 
			appeq/entov/wc, sboeq &  9.78 & 13.69  & 8.15 & 10.99\\\hline
			appeq/entov/wc, sboov &  10.48 & 15.37 & 8.66 & 12.54\\\hline 
			appeq/entov/wc, sboisa &  14.67 & 23.88  & 11.95 & 19.90 \\\hline 
			\multicolumn{5}{|l|}{} \\\hline
			\multicolumn{5}{|l|}{MTOR-HMN/MTOR-NLP} \\\hline
			&  \multicolumn{2}{c|}{nodes} & \multicolumn{2}{|c|}{edges} \\\hline
			& prec & rec & prec & rec \\\hline
			nmeq, sboeq  &  6.31 & 13.25  & 5.84 & 8.67 \\\hline
			nmeq, sboov  &  7.26 & 17.40  & 6.67 & 11.48 \\\hline 
			nmeq, sboisa  &  9.85 & 27.73  & 8.88 & 17.50\\\hline 
			nmeq/wc, sboeq &  9.83 & 40.19  & 9.21 & 31.14 \\\hline
			nmeq/wc, sboov &  10.82 & 44.34  & 10.08 & 34.43\\\hline 
			nmeq/wc, sboisa &  14.48 & 58.58  & 13.30 & 46.68 \\\hline 
			appeq, sboeq  & 6.56 & 14.04  & 6.07 & 9.24 \\\hline
			appeq, sboov  &  7.53 & 18.69  & 6.92 & 12.37 \\\hline 
			appeq, sboisa  &  10.39 & 30.35  & 9.35 & 19.41\\\hline 
			appeq/wc, sboeq &  10.27 & 40.83  & 9.63 & 31.62\\\hline
			appeq/wc, sboov &  11.28 & 45.53  & 10.52 & 35.33\\\hline 
			appeq/wc, sboisa &  15.24 & 60.85  & 13.98 & 48.43 \\\hline 
			appeq/enteq, sboeq  &  9.33 & 18.44  & 8.64 & 12.21 \\\hline
			appeq/enteq, sboov  &  11.06 & 23.63  & 10.16 & 15.71 \\\hline 
			appeq/enteq, sboisa  &  15.94 & 37.22  & 14.28 & 24.50\\\hline 
			appeq/enteq/wc, sboeq & 21.40 & 49.73  & 20.11 & 40.58\\\hline
			appeq/enteq/wc, sboov & 23.59 & 55.66  & 22.06 & 45.95\\\hline 
			appeq/enteq/wc, sboisa & 32.88 & 75.33 & 30.18 & 65.04 \\\hline
			 
			appeq/entov, sboeq  & 20.18 & 44.44  & 18.90 & 34.88 \\\hline
			appeq/entov, sboov  &  22.35 & 50.32 & 20.83 & 39.97 \\\hline 
			appeq/entov, sboisa  &  31.34 & 69.85  & 28.65 & 57.51\\\hline 
			appeq/entov/wc, sboeq & 21.40 & 49.73  & 20.11 & 40.58\\\hline
			appeq/entov/wc, sboov &  23.59 & 55.66  & 22.06 & 45.95\\\hline 
			appeq/entov/wc, sboisa & 32.88 & 75.33  & 30.18 & 65.04 \\\hline 
		\end{tabular}
	\end{center}
	\caption{Results of matching MTOR-ANN and MTOR-NLP with MTOR-HMN.
		Results are always precision/recall.}
	\label{t:network}
\end{footnotesize}
\end{table}

\paragraph{Caveats}
There are number of issues that need to be taken into account when analyzing these
results. For instance, SBO/GO term annotation for MTOR-HMN is not perfect, as can
be seen from the large number of conversion operations. Similarly, Entrez Gene id 
normalization has its problems, especially when dealing with complex species. Lastly,
reaction signature overlap does not count reactions with multiple reaction types
as separate. We are currently working on dealing with each of these issues.
Some will arguably improve performance, others decrease
precision and recall numbers. We are confident though that the general trends in the
results will uphold.

\section{Discussion}
The last section quantitatively demonstrated differences
between extraction and curation. Curation involves 
processes such as annotation, selection and, in particular, 
composition (of subgraphs into a large graph). 
The next paragraphs summarize the most important problems.

\paragraph{Species Normalization} There has been a lot 
of work on this topic  \cite{van2013large,wei2015simconcept,sohn2008abbreviation,dougan2014finding,hakenberg2011gnat}
provide impressive performance. But there is the problem of
how to use the information provided by tools such as GNAT.
GNAT, for instance, returns hypotheses of possible identifiers.
It is then up to subsequent systems to use this information
and reject certain hypotheses based on other information in the text.

\paragraph{Complex formation} Identification of complexes is missing from NLP extraction systems. 
To the best of our knowledge, there is very little work on extraction of complexes and their participants from text 
(except generally in terms of Named Entity Recognition). However, complexes are extremely important
for the mTOR pathway. For a large part the pathway consists of complexes that form and subsequently modify other reactions.
Not being able to extract such information is a significant disadvantage for automated systems.

\paragraph{Composition of pathways} The NLP system produces 
pathway maps that consist of scattered reactions without integrating them into one. The human
map on the other hand is all about a single network of reactions. Composition is a combinatorial problem 
constrained by cues in the Natural Language as well as biology.
This paper proposed a number of matching  strategies. These strategies are not
only useful for measuring the state-of-the-art. For instance, matching 
of species based on Entrez Gene normalization could be useful in pathway
composition.

\paragraph{Understanding levels of detail of representation} A fundamental problem 
in pathway curation is that information can be represented on different levels of specificity.
For instance, it might be sufficient to capture phosphorylation instead of capturing 
the exact sites or the number of phosphoryl groups added. Often human modelers make
various abstractions and conceptualizations of the same underlying biological process.
Final pathway maps are affected by prior knowledge of the curator and this shapes 
the pathway that a human produces. The problem then becomes how to build machines that can
extract knowledge on various levels of abstraction. 

It is important to realize that these issues are not just a problem of more data or more precise annotation. 
Current NLP systems are good at classifying strings and their relations but they have no notion of the underlying processes
(in this case the biological processes involved). The learning signal of NLP systems is annotated text
and it is not the human-curated biological model. The human as an expert in Systems Biology reading the text
will pick out relevant detail and try to build a consistent overall model based on the information
in the various texts. The NLP system relies on information detected in the text without any
actual notion of what the text actually means, i.e. without building an internal model and
integrating it with prior information.

\section{Conclusion}
To the best of our knowledge, this paper is the first to evaluate automated pathway extraction systems
by measuring the difference between automated systems and human curation.
We believe this kind of analysis is crucial to make progress towards the ultimate goal of 
complete automation of pathway curation. The contribution of this paper is twofold: 1)
we propose a number of measures that can be used to quantify the state-of-the-art;
2) we identify a number of areas where progress can improve the state-of-the-art
measurably.

This paper is part of a larger trend in NLP to move from event extraction to 
knowledge base creation \cite{kim2015extending} and construction of biologically
relevant networks \cite{rinaldi2016biocreative}. 
It is therefore perfectly aligned with people trying to automatically build mechanistic 
dynamic pathway models \cite{bigmechanism} that could ultimately have a big scientific impact \cite{kitano2016nobel}.

\bibliographystyle{acl2016}
\bibliography{acl2016}

\end{document}